\documentclass[11pt]{article}
\usepackage[margin=1in]{geometry}
\usepackage[round]{natbib}
\usepackage{amsthm}
\usepackage{amsfonts}
\usepackage{microtype}
\usepackage{array}
\usepackage{booktabs}
\usepackage{algorithm}
\usepackage{algpseudocode}
\usepackage{textcomp}
\usepackage{url}
\usepackage{graphicx}
\usepackage{amsmath, amssymb}
\usepackage{mathtools}
\usepackage{enumitem}
\usepackage{multirow}
\usepackage{wrapfig}
\usepackage{xcolor}
\definecolor{newc}{RGB}{0,105,125}
\newif\ifnewcolor \newcolortrue
\newcommand{\new}[1]{\ifnewcolor{\color{newc}#1}\else#1\fi}
\newenvironment{newpart}{\ifnewcolor\color{newc}\fi}{}
\newcommand{\newtable}{\ifnewcolor\color{newc}\fi}
\providecommand{\tablesize}{\small}

\theoremstyle{plain}
\newtheorem{theorem}{Theorem}[section]
\newtheorem{lemma}[theorem]{Lemma}

\theoremstyle{definition}
\newtheorem{definition}[theorem]{Definition}

\theoremstyle{remark}

\newcommand{\diagramabbrev}{MGP}

\newcommand{\rephine}{RePHINE}
\newcommand{\reph}{\mathcal{R}}

\newcommand{\abs}[1]{\left\lvert #1 \right\rvert}

\newcommand{\fullparlines}{\setlength{\parfillskip}{0pt plus 0.75\linewidth}\setlength{\emergencystretch}{1.5em}}
\newcommand{\plainparlines}{\setlength{\parfillskip}{0pt plus 1fil}}
\AtBeginDocument{\fullparlines}

\newcolorfalse
\newcommand{\codeurl}{\url{https://github.com/JorgeLuizFranco/message-passing-graph-persistence}}
\usepackage{hyperref}
\hypersetup{colorlinks,citecolor=blue!50!black,linkcolor=blue!50!black,urlcolor=blue!50!black}

\title{Message-passing graph persistence}
\author{Jorge Luiz Franco\textsuperscript{1,2*}\qquad Arthur Queiroz Moura\textsuperscript{1*}\qquad Moacir Ponti\textsuperscript{1,3}\qquad Amauri Souza\textsuperscript{2,4}\\[6pt]
\small \textsuperscript{1}Institute of Mathematics and Computer Sciences (ICMC), University of S\~ao Paulo, S\~ao Carlos, Brazil\\
\small \textsuperscript{2}Avra\quad
\textsuperscript{3}Mercado Libre\quad
\textsuperscript{4}Federal Institute of Cear\'a (IFCE), Fortaleza, Brazil}
\date{}

\begin{document}
\maketitle
{\renewcommand{\thefootnote}{}\footnotetext{\textsuperscript{*}Equal contribution. Correspondence to: \texttt{jorge.luiz@usp.br}}}
\begin{abstract}
\new{Message-passing graph neural networks (GNNs) are limited in expressive power by the one-dimensional Weisfeiler--Leman test. Persistent homology (PH) provides topological descriptors that can complement GNN representations, but existing approaches typically compute persistence separately from the representation learning process. In this paper, we introduce Message-Passing Graph Persistence (MGP), a family of topological descriptors that incorporates message passing into graph filtrations. At each filtration step, node embeddings are updated using the current subgraph and associated with persistence events. These embeddings also determine the representatives of connected components during persistence computation. We prove that MGP is invariant under graph isomorphism and that it is strictly more expressive than the PH-based descriptors it extends. We evaluate MGP on an expressivity benchmark and on several graph classification and regression datasets. Experiments show that MGP outperforms other PH-based descriptors on most of these datasets.}

\end{abstract}


\section{Introduction}
Over the past decade, \emph{graph neural networks} (GNNs) \citep{Scarselli05,Bruna14,gilmer_gnn} have become the \emph{de facto} approach for predictive tasks on graph-structured data. At each layer, typical GNNs sequentially combine local message-passing mechanisms, non-linear activation functions, and (possibly) pooling operations to obtain refined graph representations. Despite their empirical success, the expressive power of the most widely adopted GNNs is theoretically bounded by the one-dimensional Weisfeiler--Leman (1-WL) test \citep{weisfeiler1968reduction} for graph isomorphism \citep{gin,Morris2019}. As a result, these models fail to distinguish even simple families of graphs (e.g., regular graphs) \citep{gin} or to decide basic graph properties (e.g., number of connected components) \citep{Garg2020}.

To overcome this limitation, \emph{topological data analysis} (TDA), and in particular \emph{persistent homology} (PH), offers a principled framework for extracting multi-scale topological features from data \citep{chazal2021introduction, coskunuzer2024topologicalmethodsmachinelearning,chen2021topological}. PH tracks the appearance and disappearance of topological invariants across a filtration --- a nested sequence of subspaces. The resulting information is encoded in a persistence diagram, which summarizes the lifespan of each feature as a multiset of birth--death pairs.

Several approaches \citep{rephine,horn2022topological,pmlr-v235-verma24a,ji2025graphpersistencegoesspectral} have sought to integrate PH-based descriptors into graph learning models (e.g., GNNs and kernel methods) to improve expressivity and predictive performance. For instance, \citet{horn2022topological} introduced TOGL, a general framework for integrating topological features derived from PH into GNN layers. \citet{rephine} analyzed the representational ability of color-based PH schemes, providing a complete characterization of the power of $0$-dimensional PH methods. Building on this, they introduced RePHINE as a strictly more powerful descriptor for graph classification tasks.
More recently, SpectRe \citep{ji2025graphpersistencegoesspectral} proposed enhancing RePHINE diagrams by augmenting them with spectral information, namely eigenvalues computed from graphs along a filtration.

Despite these advances, existing PH-based descriptors are computed in a stage that remains largely decoupled from the iterative representation learning process. Although filtration functions may be learned or derived from node features, the persistence computation itself does not dynamically interact with message passing across scales. In addition, standard GNN embeddings rely on local message passing over the full graph and do not explicitly leverage multi-scale information.

In this paper, we introduce \textit{Message-Passing Graph Persistence} (MGP), a novel family of topological descriptors that embeds message-passing mechanisms directly into the PH filtration pipeline. At each filtration step, node embeddings are updated via a message-passing GNN layer, and these evolving ``colors'' (or embeddings) are attached to persistence events and govern the representative selected at component merges.

\section{Background}

We represent an attributed (or colored) undirected graph as a tuple $G=(V,E,x)$, where $V$ is a set of vertices, $E$ is a set of unordered pairs of vertices, called edges, and $x:V\cup E\to\mathcal{X}$ assigns a feature (or color) to each vertex and edge. The color space is the disjoint union \(\mathcal{X}=\mathcal{X}_V\bigsqcup\mathcal{X}_E\), so vertex and edge colors remain different types even when their numerical encodings coincide. We write \(x_v=x(v)\) and \(x_{u,v}=x(\{u,v\})\). For simplicity of the theoretical presentation, and following RePHINE~\citep{rephine}, we assume that each edge color is determined symmetrically by the colors of its endpoints: \(x_{u,v}\) depends only on \(x_u,x_v\) and is unchanged when the endpoints are exchanged.

Two graphs $G=(V,E,x)$ and $G'=(V',E',x')$ are said to be \emph{isomorphic} if there is a bijection $\sigma:V\to V'$ such that $\{u,v\}\in E$ if and only if $\{\sigma(u),\sigma(v)\}\in E'$, and $x'_{\sigma(v)}=x_v$ for every $v\in V$. Since both graphs use the same symmetric endpoint-color rule, corresponding edges also satisfy $x'_{\sigma(u),\sigma(v)}=x_{u,v}$. We represent the set of neighbors of a vertex $v$ by $\mathcal{N}_v(G)=\{u:\{u,v\}\in E\}$ and use $\{\!\!\{\cdot\}\!\!\}$ to denote multisets.

\subsection{Persistent homology (PH) on graphs}

The core idea of PH is to keep track of the appearance and disappearance of topological features (e.g., connected components, loops) in a filtration.
A \textit{filtration} of a graph $G$ is a non-decreasing indexed family $\{G_\alpha\}_{\alpha \in \mathbb{R}}$ of subgraphs of $G$, i.e., $ G_{\alpha} \subseteq G_{\alpha'}$ if $\alpha < \alpha'$, with $G_\infty=G$.
Filtrations are induced by filtering functions (also known as filtration functions) $f: V \cup E \rightarrow \mathbb{R}$, which assign a number to each vertex and edge such that, for every $\{u,v\} \in E$, $f(\{u, v\}) \geq \max\{f(u), f(v)\}$. This condition ensures that no edge appears in a filtration before both of its endpoints have been included.
For each $\alpha \in \mathbb{R}$, we define $G^f_\alpha = f^{-1}((-\infty, \alpha])$, that is,
\[
    G^f_\alpha = \{v \in V : f(v) \leq \alpha \}  \bigcup \{ \{u, v\} \in E : f(\{u, v\}) \leq \alpha \}
\]
and denote the \textit{sublevel filtration} of $G$ induced by $f$ as $\{G^f_\alpha\}_{\alpha \in \mathbb{R}}$.
Note that the subgraphs in a filtration only differ at critical values $\alpha \in \mathrm{Im}{f}$, also known as filtration steps. Thus, we can reduce the filtration to the finite sequence of subgraphs at these critical values. With this in mind, we can capture the persistence of topological features in a filtration, i.e., how these features change over time. In particular, if a topological feature first appears in $G^f_{\alpha_b}$ and disappears in $G^f_{\alpha_d}$, then we encode its persistence as a pair $(\alpha_b, \alpha_d)$; if a feature does not disappear, then its persistence is $(\alpha_b, \infty)$.
The collection of all pairs forms a multiset called a \textit{persistence diagram}.
We use $\mathcal{D}_{i}(G, f)$ to denote the persistence diagram for the $i$-th dimensional topological features of $G$ induced by $f$. In particular, here we consider only the 0- and 1-dimensional topological features, i.e., \(i=0\) or \(i=1\).

\noindent \textbf{Color-based filtration functions.} Following prior work~\citep{rephine,ji2025graphpersistencegoesspectral}, we use colors to define filtration functions in an isomorphism-compatible way. Recall that \(\mathcal{X}=\mathcal{X}_V\bigsqcup\mathcal{X}_E\), so vertex and edge colors remain distinct even when represented by the same numerical vector.

A \textit{color filtering function} is a map \(f:\mathcal{X}\to\mathbb{R}_{\geq 0}\). It induces functions on the simplices of an attributed graph \(G=(V,E,x)\):
\begin{enumerate}
    \item the \textit{vertex-based filtration function} \(F_V^f:V\cup E\to\mathbb{R}_{\geq 0}\), defined by
    \begin{align*}
        F_V^f(v)&:=f(x_v),\\
        F_V^f(\{u,v\})&:=\max\{f(x_u),f(x_v)\};
    \end{align*}
    \item the \textit{edge-based filtration function} \(F_E^f:V\cup E\to\mathbb{R}_{\geq 0}\), defined by
    \[
        F_E^f(v):=0,\qquad
        F_E^f(\{u,v\}):=f(x_{u,v}).
    \]
\end{enumerate}
Both functions satisfy the filtration condition. In the first filtration, an edge appears as soon as both endpoints have appeared. In the second, all vertices are present initially and edges are subsequently added according to their colors. The distinction between the color filtering function \(f\) and the induced simplex-wise functions \(F_V^f,F_E^f\) will be useful below.

\noindent \textbf{Persistence diagrams.} Let \(G\) be an attributed graph and let \(f\) be a color filtering function. We define the \textit{PH descriptor} induced by \(f\) as
\[
\mathrm{PH}(G,f)=\bigl(\mathrm{PH}^V(G,f),\mathrm{PH}^E(G,f)\bigr),
\]
where
\begin{align}
\mathrm{PH}^V(G,f)&:=\mathcal{D}_0(G,F_V^f)\sqcup\mathcal{D}_1(G,F_V^f),\\
\mathrm{PH}^E(G,f)&:=\mathcal{D}_0(G,F_E^f)\sqcup\mathcal{D}_1(G,F_E^f),
\end{align}
and \(\sqcup\) denotes multiset disjoint union. The two components are kept separate in the descriptor, so the homological dimension and filtration type remain identifiable.

\noindent \textbf{\rephine\ diagram.} RePHINE \citep{rephine} uses an edge-color filtering function \(f:\mathcal{X}_E\to\mathbb{R}_{\geq 0}\) and an auxiliary vertex-color function \(g:\mathcal{X}_V\to\mathbb{R}\). Its descriptor is \(\reph(G,f,g)=\reph_0(G,f,g)\sqcup\reph_1(G,f,g)\), where \(\sqcup\) denotes multiset disjoint union and:
\begin{enumerate}
    \item \(\reph_0(G,f,g)\) contains one tuple \((0,\alpha_{d_v},\lambda_v,\gamma_v)\) for each vertex \(v\). Here \(\alpha_{d_v}\) is the death time of the component representative associated with \(v\) under the edge filtration induced by \(f\), \(\lambda_v=g(x_v)\), and
    \[
        \gamma_v=\min_{u\in\mathcal{N}_v(G)}f(x_{u,v}),
    \]
    using the convention \(\min\varnothing=0\).
    When components merge, RePHINE kills the representative with the largest \(\lambda_v\); ties are resolved by killing the representative with the smallest \(\gamma_v\), and any remaining tie may be resolved arbitrarily.
    \item \(\reph_1(G,f,g)\) contains \((1,f(x_{u,v}),0,0)\) for each edge \(\{u,v\}\) that creates a cycle.
\end{enumerate}
Thus, RePHINE is built on an edge filtration, while vertex colors enter through the auxiliary feature \(g\).

\subsection{Graph neural networks (GNNs)}
Most popular GNNs can be described using the message-passing paradigm~\citep{gilmer_gnn,gin}. Within this framework, a GNN initializes each node's embedding with its original features, i.e., $h^{(0)}_v = x_v$ for all nodes $ v \in V$. Then, at each layer $\ell$, each node $v$ gathers messages from its neighbors $u \in \mathcal{N}_v(G)$, compiling them into an aggregated message $m^{(\ell)}_v$:
\begin{equation*}
    m_v^{(\ell)} = \textsc{Aggregate}_\ell\left(\{\!\!\{ h^{(\ell-1)}_u  : u \in \mathcal{N}_v(G)\}\!\!\}  \right),  \quad \quad
\end{equation*}
where $\textsc{Aggregate}_\ell$ is an arbitrary function on multisets, i.e., it is order-invariant. Subsequently, each node uses the (so-called) $\textsc{Update}$ function (e.g., a feedforward neural network) to refresh its embedding in light of the aggregated message:
\begin{equation*}
    h_v^{(\ell)} = \textsc{Update}_\ell\left( m_v^{(\ell)}, h_v^{(\ell-1)}  \right).
\end{equation*}

\subsection{Weisfeiler-Leman and the expressive power of GNNs}
The Weisfeiler-Leman (WL) graph isomorphism test \citep{weisfeiler1968reduction} iteratively refines a labeling of vertices by collecting neighbor labels into a multiset and hashing it together with the vertex's current label. It applies the following iterative procedure:
\begin{enumerate}[leftmargin=1cm, align=left]
\item[\emph{Initialization}:] The colors of all nodes in $G$ are initialized using the initial node features: $\forall v \in V,  h^{0}(v) = x_v$. If node features are not available, all nodes receive identical colors;
\item[\emph{Refinement}:] At step $\ell$, the colors of all nodes are refined using a hash (injective) function: for all $v \in V$, we apply $h^{\ell+1}(v) = \textsc{Hash}(h^{\ell}(v), \{\!\!\{h^{\ell}(u): u \in \mathcal{N}_v(G)\}\!\!\})$;
\item[\emph{Termination}:] The test is carried out for two graphs in parallel and stops when the multisets of corresponding colors diverge, returning non-isomorphic. If the algorithm runs until the number of different colors stops increasing, the test is deemed inconclusive.
\end{enumerate}

The 1-WL test mirrors the message-passing structure of GNNs and thus provides an upper bound on their distinguishing power. In particular, since 1-WL aggregates neighbor labels as multisets and updates node representations via an injective hashing scheme, any GNN operating under the message-passing paradigm is at most as expressive as 1-WL \citep{gin}. Moreover, when both the $\textsc{Aggregate}$ and $\textsc{Update}$ functions in a GNN are injective (i.e., incur no loss of information), the resulting message-passing procedure coincides with the refinement step of the 1-WL test. In this setting, the GNN achieves the same discriminative power as the 1-WL test. This injective idealization is standard in GNN expressivity theory: GIN uses injective aggregation and update maps to characterize maximal message-passing expressivity and gives sum/MLP constructions for the required multiset functions~\citep{gin}.

\section{Related work}

Early efforts to combine PH with graph learning predate end-to-end
architectures. \citet{rieck2019persistent} propose the
persistent Weisfeiler--Leman procedure (P-WL), which enriches the WL
relabeling scheme with topological information: distances between
propagated label multisets are used to assign edge weights, turning an
unweighted labeled graph into a metric one whose sublevel filtration
yields persistence diagrams. The resulting features generalize WL
subtree features, and the observed gains are largely attributable to
cycle information, which subtree patterns cannot capture. However,
P-WL operates as a fixed feature-extraction scheme: the label
propagation and the persistence computation are performed once,
prior to and independently of any downstream learning.

Building on the idea of task-adapted topology, \citet{hofer2020graphfiltration} introduce graph filtration learning
(GFL), in which vertex filtration values are produced by a learnable
function of node features and gradients are propagated through the
persistence computation, allowing the filtration to be optimized
end-to-end. While this renders the filtration adaptive, the node
representations that induce it are computed on the full graph before
persistence is evaluated; the diagram computation itself remains a
post-hoc summary that does not interact with message passing along
the filtration.

\new{PEGN \citep{zhao2020pegn} uses PH to reweight GNN messages for node classification. Persistent message passing \citep{strathmann2021persistent}, despite its name, does not use PH.}

Recently, PH has been used to boost the expressive power of GNNs. \citet{horn2022topological} introduce TOGL --- a general approach for incorporating topological features from PH into GNN layers. In particular, TOGL leverages node embeddings at each layer of a GNN by using vertex-color filtrations. More precisely, the 0-dim individual persistence tuples are vectorized using Multilayer Perceptrons (MLPs) and appended to the corresponding node features at each layer. For 1-dim tuples, TOGL applies DeepSets~\citep{DeepSets} to get a graph-level vector that runs through the final fully-connected layers of the GNN.

\rephine~\citep{rephine} uses independent vertex-color and edge-color filtering functions to obtain more expressive persistence diagrams. More specifically, RePHINE first computes persistence diagrams from a filtration induced by edge colors. Each tuple of the diagram is then augmented based on the vertex colors and the local edge-color information around each vertex. RePHINE diagrams are vectorized using DeepSets~\citep{DeepSets} and combined with graph-level GNN embeddings in the final classifier. \new{SpectRe~\citep{ji2025graphpersistencegoesspectral} augments RePHINE tuples with the Laplacian spectrum of the component associated with each event.}

PersLay \citep{carriere2020perslay} was introduced as a differentiable, permutation-invariant layer dedicated to processing persistence diagrams. Each diagram is represented as a permutation-invariant operation (e.g., sum) of kernel functions centered on its points, with both the kernel parameters and the weighting functions learned end-to-end. By integrating PersLay into a GNN architecture, topological summaries can be injected at arbitrary depths of the network, enabling the model to adaptively emphasize specific homological features (e.g., long-lived cycles) during training.

\new{\citet{ballester2024expressivity} relate the expressivity of PH to the Weisfeiler--Leman hierarchy, while \citet{eitan2025topological} identify invariants that higher-order message passing cannot capture.}

In summary, existing approaches either extract topological features
prior to learning~\citep{rieck2019persistent}, learn the filtration but
compute persistence as a separate downstream stage~\citep{hofer2020graphfiltration,
horn2022topological, carriere2020perslay}, and/or enrich the persistence
tuples with  color information~\citep{rephine,
ji2025graphpersistencegoesspectral}. None of them performs message passing on the
intermediate filtration subgraphs. For the integration with GNN
architectures, we adopt the framework used by \rephine~\citep{rephine}.

\section{Method}\label{sec:rephine-reinterpretation}
Instead of viewing the \rephine\ diagram as arising from an edge-based filtration augmented with the additional neighborhood descriptor \(\gamma_v\), we can adopt an alternative perspective: modify the filtration so that the first incident-edge value becomes the birth time of each vertex.

Let \(f:\mathcal{X}_E\to\mathbb{R}_{\geq 0}\) be an edge-color filtering function. We adopt the convention that the minimum of an empty set of incident-edge values is zero. Define the induced \textit{structure-aware edge filtration function} \(F_S^f:V\cup E\to\mathbb{R}_{\geq 0}\) by
\begin{align}
F_S^f(v)&=\min_{u\in\mathcal{N}_v(G)}f(x_{u,v}),\label{eq:structural-vertex}\\
F_S^f(\{u,v\})&=f(x_{u,v}).\label{eq:structural-edge}
\end{align}
This is a valid filtration function because \(F_S^f(u),F_S^f(v)\leq F_S^f(\{u,v\})\) for every edge \(\{u,v\}\). A non-isolated vertex therefore appears exactly when its first incident edge appears.

The underlying subgraph sequence differs from RePHINE's edge filtration, in which every vertex is present at time zero. To compare the resulting tuples, we use the following conventions. When vertices and incident edges have the same filtration value, the vertices are inserted first; zero-persistence vertex--edge pairs are retained; and component deaths are associated with vertices using the RePHINE representative rule. Under these conventions, each RePHINE zero-dimensional tuple
\[
(0,\alpha_{d_v},\lambda_v,\gamma_v)
\]
corresponds to the tuple
\[
(\gamma_v,\alpha_{d_v},\lambda_v)
\]
in the structure-aware filtration: \(\gamma_v\) becomes the vertex birth time, while the death time and associated vertex feature are unchanged. Conversely, the RePHINE tuple is recovered by restoring the fixed birth coordinate zero and moving the structural birth coordinate back to the fourth entry. Likewise, a missing-hole tuple \((1,d,0,0)\) carries the same information as the one-dimensional persistence tuple \((d,\infty)\). This is only a correspondence between individual tuples; it does not assert that the two filtrations have the same subgraph sequence.

This reinterpretation removes the redundant component \(b_v\equiv0\) and places \(\gamma_v\) in the role of a birth coordinate. It follows the vertex-wise pairing convention used by RePHINE and SpectRe \citep{rephine,ji2025graphpersistencegoesspectral}.

\subsection{Persistent homology meets message-passing GNNs}
\label{sec:mgp-def}
We now define MGP, which replaces \(\lambda_v\) by embeddings computed along the filtration.

\noindent \textbf{Overview.} Given a structure-aware edge filtration \(\{G^{F_S^f}_{\alpha_i}\}_{i=1}^n\), an embedding function \(\Psi\), and message-passing functions, we first transform the initial node features using \(\Psi\). We then iterate through the filtration steps. After all edges with value \(\alpha_i\) are added, we perform one round of message passing on the current subgraph. The zero-dimensional tuples have the form \((\alpha_{b_v},\alpha_{d_v},h_v)\), where \(b_v\) and \(d_v\) are the birth and death indices of the representative associated with \(v\), and \(h_v\) is its embedding at death, or at the final filtration step if it survives.

\begin{definition}[MGP diagram]\label{def:mgp}
Let \(\alpha_1<\cdots<\alpha_n\) be the critical values of the structure-aware filtration of \(G=(V,E,x)\) induced by \(F_S^f\). Given \(\Theta=(\Psi,\psi,\phi,\bigoplus)\), define
\[
\operatorname{MGP}(G,f,\Theta)
=\operatorname{MGP}_0(G,f,\Theta)\sqcup\operatorname{MGP}_1(G,f),
\]
where \(\sqcup\) denotes multiset disjoint union, so the summand records the homological dimension, and:
\begin{enumerate}[left=8pt]
\item Each element of \(\operatorname{MGP}_0(G,f,\Theta)\) is a tuple \((\alpha_{b_v},\alpha_{d_v},h_v)\) associated with a vertex \(v\). Here \(b_v\) is the first filtration index at which \(v\) appears, and \(d_v\) is the index at which its representative dies. If it never dies, we set \(d_v=\infty\) and interpret \(\alpha_{d_v}=\infty\). The embeddings belong to a linearly ordered space \((\mathcal{H},\leq)\), for example \(\mathbb{R}^k\) under lexicographic ordering, and are computed recursively by
\begin{align*}
h_v^{(b_v-1)}&=\Psi(x_v),\\
m_v^{(\ell)}&=\bigoplus\!\left\{\!\!\left\{
\begin{aligned}
&(h_u^{(\ell-1)},h_v^{(\ell-1)},\phi(x_{u,v})):\\[-2pt]
&\hspace{18mm}u\in\mathcal{N}_v(G^{F_S^f}_{\alpha_\ell})
\end{aligned}
\right\}\!\!\right\},\\
h_v^{(\ell)}&=\psi\!\left(h_v^{(\ell-1)},m_v^{(\ell)}\right),
\qquad b_v\leq \ell\leq n,\\
h_v&=\begin{cases}
h_v^{(d_v)},&d_v<\infty,\\
h_v^{(n)},&d_v=\infty.
\end{cases}
\end{align*}
Here \(\Psi:\mathcal{X}_V\to\mathcal{H}\) initializes node colors, \(\phi\) embeds the edge colors \(x_{u,v}\), \(\bigoplus\) is permutation-invariant, and \(\psi\) is an update function.

A component has one alive representative. When several components merge, representatives with the largest terminal embeddings \(h_v^{(n)}\) die until one remains. If terminal embeddings tie, any tied representative may be selected; Theorem~\ref{thm:isomorphism-invariant} gives a condition under which this choice does not change the descriptor.

\item \(\operatorname{MGP}_1(G,f)=\mathcal{D}_1(G,F_S^f)\), that is, each edge \(\{u,v\}\) that creates a cycle contributes \((f(x_{u,v}),\infty)\).
\end{enumerate}
\end{definition}

\begin{newpart}
\noindent \textbf{Step schedule.} For learned filtrations, the number of critical values can be of the order of the number of edges, requiring the same order of message-passing rounds. To limit this cost, we group the critical values of each graph into \(T\) quantile steps. An edge \(\{u,v\}\) of a graph with edge set \(E\) is assigned to step \(\lceil T\,\abs{\{e'\in E: f(x_{e'})\leq f(x_{u,v})\}}/\abs{E}\rceil\in\{1,\dots,T\}\). Thus, edges with equal filtration values belong to the same step, and a vertex is born at the first step containing one of its incident edges. We apply Definition~\ref{def:mgp} and Algorithm~\ref{alg:mgp-algo} using the steps \(1,\dots,T\) instead of the indices of the critical values, while keeping the original filtration values in the tuples. The case \(T=\infty\) corresponds to the distinct-value schedule. Since the grouping depends only on the multiset of filtration values, isomorphic graphs have isomorphic subgraphs at corresponding steps. Therefore, the invariance argument still holds. We treat \(T\) as a hyperparameter. Within each step, edges are processed in increasing order of \(f\). The death value is given by the filtration value of the merging edge, while the corresponding embedding is computed at the end of the step.

\noindent \textbf{Ties.} When representatives have identical terminal embeddings, we keep the one with the lower vertex index. Under partial injectivity, identical terminal embeddings correspond to identical computation trees, and any choice produces the same persistence tuples (Theorem~\ref{thm:isomorphism-invariant}).

The implementation is described in Algorithm~\ref{alg:mgp-algo}, Appendix~\ref{app:runtime}.
\end{newpart}

\begin{figure}[t]
\centering
\includegraphics[width=0.85\textwidth]{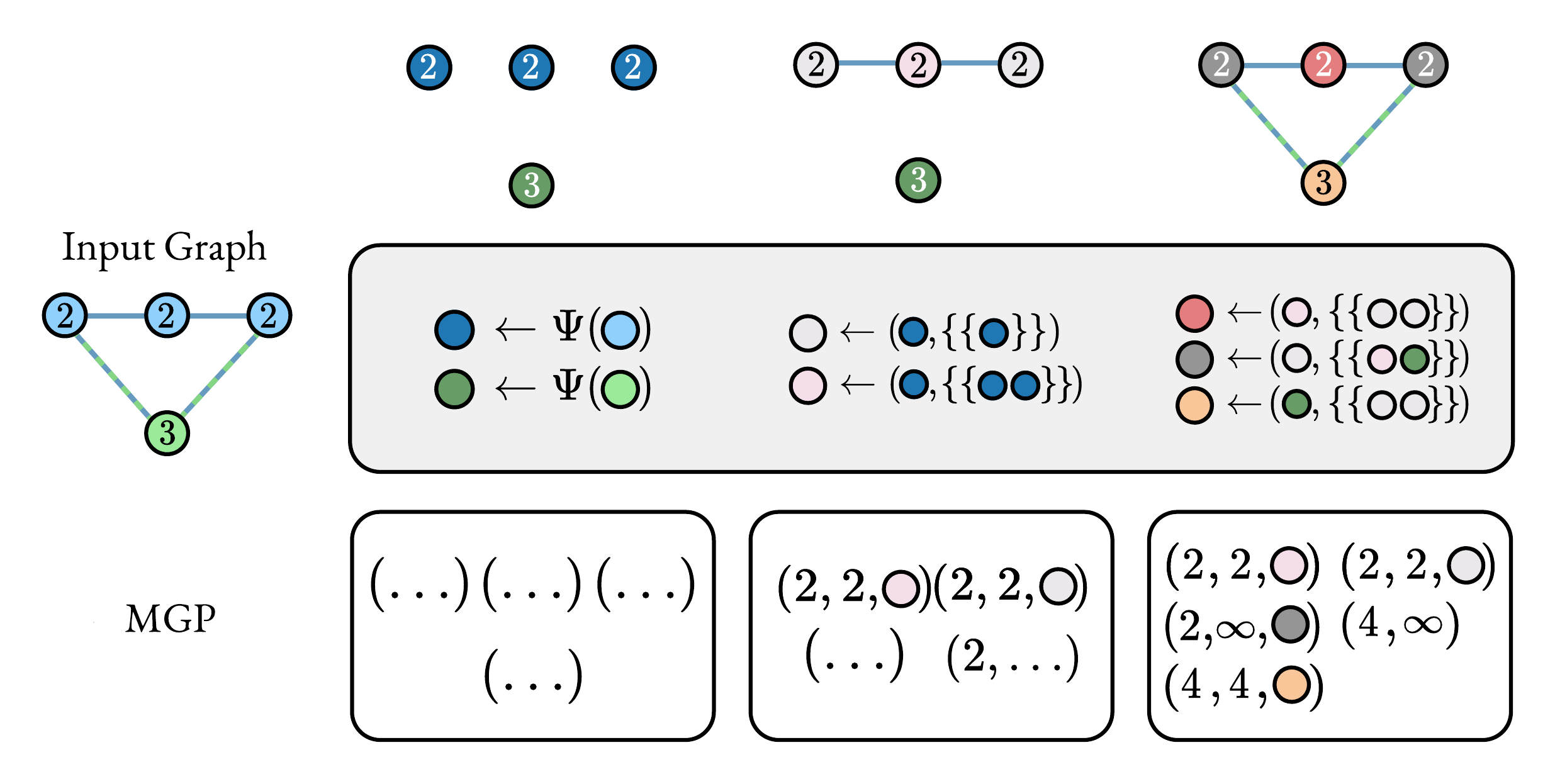}
\caption{Worked computation of the MGP diagram. Vertices are born according to the structure-aware edge filtration, and their embeddings are updated by message passing on each filtration subgraph. Zero-dimensional tuples record birth, death, and the corresponding death-step embedding, while cycle-creating edges contribute one-dimensional tuples. In this example, we use the order \( \mathrm{gray} < \mathrm{red} < \mathrm{orange} \)
on the terminal embeddings.}
\label{fig:mgp-example}
\end{figure}

The tiebreaker criterion depends on \(h_v^{(n)}\), which is available only after the filtration-wise message-passing computation. The embedding computation itself does not depend on which representatives are later associated with tuples, so all embeddings can be computed first and the union--find pairing can then be run on top of them. This adds only the union--find pass and does not change the asymptotic cost of the message-passing computation. \new{Appendix~\ref{app:proofs} shows why we use the terminal embedding \(h_v^{(n)}\) instead of the embedding at the current step, and Appendix~\ref{sec:instances} describes the message functions used in the experiments and extends Theorem~\ref{thm:isomorphism-invariant} to these functions.}

\subsection{Expressivity}
\label{sec:expressivity}
The family formulation matches the filtration-dependent comparisons used for color-based PH and RePHINE \citep{rephine}.

\begin{definition}[Expressivity of descriptor families]\label{def:family-expressivity}
Let \(\mathfrak{X}\) and \(\mathfrak{Y}\) be families of graph isomorphism invariants. We write \(\mathfrak{X}\succeq\mathfrak{Y}\) if, for every pair of non-isomorphic graphs \(G,H\),
\[
\begin{aligned}
&\exists Y\in\mathfrak{Y}:Y(G)\neq Y(H)\\
&\hspace{7mm}\Longrightarrow
\exists X\in\mathfrak{X}:X(G)\neq X(H).
\end{aligned}
\]
We write \(\mathfrak{X}\succ\mathfrak{Y}\) if \(\mathfrak{X}\succeq\mathfrak{Y}\) and \(\mathfrak{Y}\not\succeq\mathfrak{X}\).
\end{definition}

To state the invariance result, we first observe that corresponding intermediate embeddings remain equal on isomorphic graphs.

\begin{lemma}[Isomorphic graphs remain isomorphic during MGP]\label{lem:isomorphic-during-execution}
Let \(G,G'\) be attributed isomorphic graphs, and let the same color filtering function \(f\) and message-passing functions \(\Theta\) be used on both. The structure-aware filtrations have the same critical values. Moreover, every isomorphism \(\sigma:G\to G'\) restricts to an isomorphism between corresponding filtration subgraphs, and \(h_v^{(\ell)}=h_{\sigma(v)}'^{(\ell)}\) at every step \(\ell\geq b_v\).
\end{lemma}

A formal proof is given in Appendix~\ref{app:proofs}. The remaining issue is an exact tie between terminal embeddings. We use the
following explicit sufficient condition, defined in Appendix~\ref{app:proofs}:
\(\Theta\) is \textit{partially injective} when equal symbolic embeddings can
arise only from identical computation trees.

\begin{theorem}[MGP is isomorphism invariant]\label{thm:isomorphism-invariant}
Let \(G,G'\) be attributed isomorphic graphs. For a fixed color filtering function \(f\) and any partially injective \(\Theta\),
\[
\operatorname{MGP}(G,f,\Theta)=\operatorname{MGP}(G',f,\Theta).
\]
\end{theorem}

Appendix~\ref{app:proofs} defines partial injectivity precisely as the
finite-set condition needed to make the representative tie rule
unambiguous.

Let \(\mathfrak{D}_{\mathrm{MGP}}\) denote the family of MGP descriptors over valid \(f\) and partially injective \(\Theta\), and let \(\mathfrak{D}_{\mathrm{RePHINE}}\) denote the family of RePHINE descriptors over valid edge and auxiliary vertex functions.

\begin{theorem}[MGP is strictly more expressive than RePHINE]\label{thm:mgp-more-expressive-rephine}
The MGP descriptor family is strictly more expressive than the RePHINE family:
\[
\mathfrak{D}_{\mathrm{MGP}}\succ\mathfrak{D}_{\mathrm{RePHINE}}.
\]
\end{theorem}

\begin{figure}[t]
\centering
\includegraphics[width=0.3\linewidth]{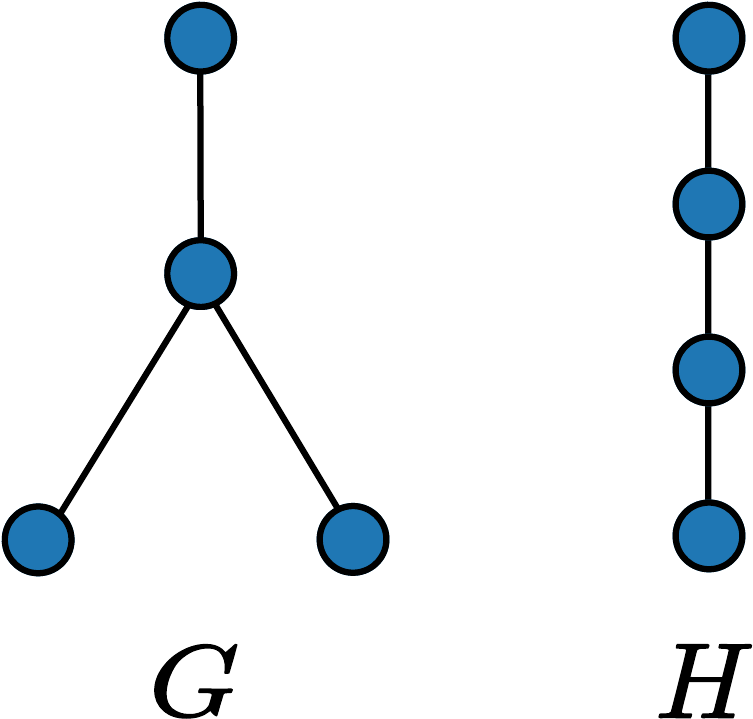}
\caption{A monochromatic four-vertex star \(G\) and path \(H\). They are indistinguishable by every element of \(\mathfrak D_{\mathrm{RePHINE}}\), whereas every element of \(\mathfrak D_{\mathrm{MGP}}\) distinguishes them.}
\label{fig:mgp-better-than-rephine}
\end{figure}

The proof is given in Appendix~\ref{app:proofs}. For strictness, consider
the monochromatic graphs in Fig.~\ref{fig:mgp-better-than-rephine}: the four-vertex star and the four-vertex path. Every RePHINE diagram contains three identical finite tuples and one identical infinite tuple on both graphs. In contrast, at the single filtration step, an injective MGP update records the degree-dependent neighborhood multiset, yielding different embedding multisets for degree patterns \(\{\!\!\{3,1,1,1\}\!\!\}\) and \(\{\!\!\{2,2,1,1\}\!\!\}\).

\subsection{GNN integration}
\label{gnn_integration}

Given any standard GNN layer (e.g., GIN~\citep{gin}, GAT~\citep{DBLP:conf/iclr/VelickovicCCRLB18}, or GCN~\citep{DBLP:conf/iclr/KipfW17}), we embed it into an MGP layer in the same way that RePHINE is combined with a GNN \citep{rephine}. \new{We use the same scheme for PH, RePHINE, SpectRe and MGP, changing only the persistence diagram.} An MGP layer has the following steps:
\begin{enumerate}
  \item \textbf{Message passing.} Apply the backbone GNN to the current node features, which are either the raw inputs or the outputs of the previous layer.
  \item \textbf{Filtrations and diagrams.} These embeddings are treated as vertex colors, and edge colors are obtained from them through the fixed symmetric endpoint rule described above. In the implementation, we use the symmetric choice \(x_{u,v}=x_u+x_v\). The layer contains multiple learnable pairs \((f_i,\Theta_i)\), each of which produces an MGP diagram \(\operatorname{MGP}(G,f_i,\Theta_i)\).
  \item \textbf{Vectorization.} The zero-dimensional tuples are processed by a DeepSet\new{ with mean pooling over the tuples of a graph}, and the one-dimensional tuples are processed by a second DeepSet\new{ with sum pooling}. Their outputs\new{, followed by a normalization layer and an MLP,} are used as graph-level embeddings for the current layer.
  \new{\item \textbf{Prediction.} We repeat Steps 1--3 at every backbone layer, obtaining one read-out per layer. The final predictor concatenates the pooled node embeddings from the last layer with the mean of the read-outs from all layers.}
\end{enumerate}

\begin{newpart}
\subsection{Complexity}
\label{sec:implementation}
A GINE-type message-passing round at step \(\ell\) has complexity \(O(\abs{E}k+\abs{V}k^2)\) for a graph with vertex set \(V\) and edge set \(E\). For \(T\) filtration steps and \(R\) rounds per step, MGP requires \(O(TR(\abs{E}k+\abs{V}k^2))\) time and \(O(T\abs{V}k)\) memory per filtration. The persistence pairing is computed using an integer union--find algorithm in \(O(\abs{E}\,\alpha(\abs{V}))\) time (Appendix~\ref{app:runtime}). Within a step, all vertices, edges, filtrations and graphs of a batch are updated in parallel on the GPU, and each round only uses sparse operations. In contrast, SpectRe computes the Laplacian spectrum of the component of every event, which costs \(O(c^3)\) for a component with \(c\) vertices and up to \(O(\abs{E}\,\abs{V}^3)\) per filtration.
\end{newpart}

\section{Results}
\label{sec:results}\label{sec:real-datasets}

\begin{table}[!b]
\centering
\newtable
\caption{Test accuracy (\%) on five TU datasets and mean absolute error on ZINC (lower is better) at the epoch of best validation score, mean \(\pm\) std over five seeds. All methods share the backbone of each dataset; best mean per column in bold.}
\label{tab:main}
\tablesize
\setlength{\tabcolsep}{3pt}
\begin{tabular}{lcccccc}
\toprule
 & \textbf{MUTAG} & \textbf{PTC\_MR} & \textbf{PROTEINS} & \textbf{NCI109} & \textbf{NCI1} & \textbf{ZINC} \\
\midrule
GNN     & 73.68 $\pm$ 12.34 & 56.00 $\pm$ 2.56 & 72.86 $\pm$ 1.20 & 79.90 $\pm$ 2.03 & 75.47 $\pm$ 3.33 & 0.162 $\pm$ 0.006 \\
PH      & \textbf{91.58 $\pm$ 2.88} & 59.43 $\pm$ 4.69 & 70.36 $\pm$ 1.32 & 78.69 $\pm$ 1.67 & 77.23 $\pm$ 2.88 & 0.160 $\pm$ 0.021 \\
RePHINE & 86.32 $\pm$ 2.88 & 60.00 $\pm$ 7.28 & 70.71 $\pm$ 2.22 & 77.68 $\pm$ 1.97 & \textbf{81.75 $\pm$ 3.09} & 0.142 $\pm$ 0.013 \\
SpectRe & 89.47 $\pm$ 3.72 & 57.71 $\pm$ 2.39 & 70.71 $\pm$ 2.48 & 79.61 $\pm$ 2.55 & 79.51 $\pm$ 1.87 & 0.157 $\pm$ 0.018 \\
MGP     & 88.42 $\pm$ 7.81 & \textbf{61.71 $\pm$ 5.19} & \textbf{73.04 $\pm$ 3.91} & \textbf{82.23 $\pm$ 1.66} & 81.36 $\pm$ 0.47 & \textbf{0.110 $\pm$ 0.011} \\
\bottomrule
\end{tabular}
\end{table}

\begin{newpart}
We evaluate on five graph-classification datasets from TUDataset \citep{Morris2020tudataset}, namely MUTAG, PTC\_MR, PROTEINS, NCI1 and NCI109, and on ZINC (12k subset, regression, mean absolute error) \citep{dwivedi2023benchmarking}. The datasets are described in Appendix~\ref{app:dataset-details}. For the TU datasets, we follow the evaluation split of \citet{rephine}, using one stratified split with 80\% training, 10\% validation and 10\% test graphs (seed 42). The same split is used for every method, while ZINC uses its public split. We report the test score at the epoch with the best validation score.

\noindent\textbf{Shared backbone.} All methods follow the architecture of Section~\ref{gnn_integration} and use the same backbone for each dataset. We first perform a random search over the backbone configuration for every method and dataset. The search includes normalization, GCN or GINE layers, jumping knowledge \citep{xu2018jk}, dropout, width, depth, random-walk structural encoding, and plateau or cosine \citep{loshchilov2017sgdr} learning-rate schedules. We sample 32 configurations for each baseline and 64 for MGP. We then select the configuration of the RePHINE run with the best validation score and use it as the shared backbone for all methods on that dataset. The selected configurations use GINE layers on every dataset (Table~\ref{tab:selected-b}). The ``GNN'' row reports the backbone without a persistence descriptor.

\noindent\textbf{Descriptors.} On PTC\_MR, all methods use the configurations selected on MUTAG. For PH, RePHINE and SpectRe, we use \(F=8\) filtrations and \(k=16\). For MGP, we perform successive halving over 96 configurations per dataset, followed by up to seven combinations of three additional options (Appendix~\ref{app:training}). We use our reimplementation of SpectRe with a reduced setting, described in Appendix~\ref{app:runtime}.

\noindent\textbf{Selection and seeds.} We select configurations using the validation score obtained with seed 0. The selected configuration is then trained with seeds 1 to 4, and we report the mean \(\pm\) standard deviation of the test scores over the five seeds.
\end{newpart}

\begin{newpart}
MGP obtains the best mean on ZINC, NCI109, PROTEINS and PTC\_MR, while RePHINE obtains the best mean on NCI1 and PH on MUTAG. On ZINC, the one-standard-deviation interval of MGP is disjoint from those of all other descriptors.
\end{newpart}

\subsection{Expressivity}

\begin{wraptable}{r}{0.43\linewidth}
\vspace{-12pt}
\centering
\newtable
\caption{Total accuracy on BREC under the reliability check, by vertex filtration function: degree, betweenness, closeness and PageRank (PR). MGP: mean over three seeds.}
\label{tab:brec-results}
\vspace{4pt}
\tablesize
\setlength{\tabcolsep}{4pt}
\begin{tabular}{lcccc}
\toprule
 & \textbf{Deg.} & \textbf{Betw.} & \textbf{Clos.} & \textbf{PR} \\
\midrule
\(\mathrm{PH}^{V}\) & 0.03 & 0.49 & 0.21 & 0.04 \\
\(\mathrm{PH}^{E}\) & 0.41 & 0.41 & 0.41 & 0.41 \\
RePHINE             & 0.41 & 0.52 & 0.50 & 0.41 \\
MGP                 & 0.41 & 0.53 & 0.49 & 0.41 \\
\bottomrule
\end{tabular}
\vspace{-8pt}
\end{wraptable}
\new{BREC \citep{wang2024brec} contains 400 pairs of non-isomorphic graphs divided into four categories. The evaluation protocol and reliability check are described in Appendix~\ref{app:brec}.}

Table~\ref{tab:brec-results} reports the accuracy of four topological descriptors (\(\mathrm{PH}^V\), \(\mathrm{PH}^E\), \rephine\ and \diagramabbrev) on BREC. Degree, Betweenness \citep{WHITE1994335betw}, Closeness \citep{FREEMAN1978215}, and PageRank \citep{Page1999ThePC} are used as vertex filtering values, while augmented Forman--Ricci curvature \citep{ricci_curvature} supplies the edge filtering values. Table~\ref{tab:brec-full} in Appendix~\ref{app:brec} gives the accuracy per category.

\new{MGP distinguishes approximately the same number of pairs as RePHINE, with differences of at most 0.01 in the total rates across seeds.}

\section{Conclusion}
\begin{newpart}
We introduced MGP, a family of persistence descriptors that performs message passing along graph filtrations and associates the resulting node embeddings with persistence tuples. We proved that MGP is invariant under graph isomorphism and strictly more expressive than RePHINE. In the experiments, MGP matches RePHINE on BREC and obtains the best mean on ZINC, NCI109, PROTEINS and PTC\_MR.
\end{newpart}

\begin{newpart}
\paragraph{Limitations and future work.}
MGP is defined on graphs: the filtration, the message passing and the persistence tuples are all over nodes and edges. We did not study the extension to higher-order domains such as simplicial and cell complexes, where message passing between cells of different dimensions was studied by \citet{bodnar2021cellular,bodnar2021weisfeiler}. This could be done by replacing a fixed graph-to-complex lifting with a learnable one \citep{franco2026difflifting}, and by replacing the GNN that runs along the filtration with a topological neural network of this kind.
\end{newpart}

\begin{newpart}
\section*{Reproducibility}
Proofs: Appendix~\ref{app:proofs}; protocol and selected configurations: Section~\ref{sec:real-datasets} and Appendix~\ref{app:training}. Code: \codeurl.
\end{newpart}

{\plainparlines
\bibliographystyle{plainnat}
\bibliography{references}}

\clearpage
\appendix

\noindent This appendix provides the full proofs of the theoretical results stated in the main text, together with additional dataset details for the experiments in Section~\ref{sec:results}\new{, the BREC protocol and implementation details}. Statements are repeated here for self-containedness.

\section{Proofs}
\label{app:proofs}

The representative rule uses the terminal embedding \(h_v^{(n)}\) because embeddings at the current step can be equal for vertices that no automorphism exchanges. Consider the path \(p-p'-q\), where \(p\) and \(p'\) have the same color and \(q\) has a different one. Suppose the edge \(\{p,p'\}\) enters the filtration before \(\{p',q\}\). After the first step, \(p\) and \(p'\) have the same active neighborhood, so \(h_p^{(1)}=h_{p'}^{(1)}\) is possible, although no automorphism maps \(p\) to \(p'\). A rule based on the current embedding would then break this tie arbitrarily. After \(\{p',q\}\) enters, \(p'\) receives a message from \(q\) and \(p\) does not, so \(h_p^{(n)}\) and \(h_{p'}^{(n)}\) can differ.

\begin{definition}[Partially injective message passing]\label{app:def:partial-injective}
For every vertex \(v\) and filtration step \(\ell\), unfold \(h_v^{(\ell)}\) into the finite symbolic computation tree obtained from the initializer \(\Psi\), the edge map \(\phi\), the multiset aggregation \(\bigoplus\), and the update map \(\psi\). We call \(\Theta=(\Psi,\psi,\phi,\bigoplus)\) \textit{partially injective} on a finite collection of filtrations if
\[
\begin{aligned}
h_v^{(\ell)}=h_w^{(j)}
\quad\Longrightarrow\quad
&\text{the corresponding symbolic}\\[-2pt]
&\text{computation trees are identical}
\end{aligned}
\]
for every pair of states generated in those filtrations.
\end{definition}

The computation tree records the initial color, the number of updates, and at every update the multiset of incoming edge-labeled computation trees. Consequently, under partial injectivity, equal terminal embeddings have the same birth step and the same embedding at every corresponding earlier step. This is precisely what is needed when the representative rule encounters an exact terminal-embedding tie. The assumption follows the standard idealization used in GNN expressivity analyses: GIN characterizes maximal message-passing expressivity through injective aggregation and update maps and realizes the required multiset maps using sum aggregation and MLPs~\citep{gin}. For practical purposes, sufficiently expressive MLP-based maps are therefore treated as injective on the finite set of states encountered in a computation; global injectivity on the entire ambient Euclidean space is not required here.

\begin{lemma}[Isomorphic graphs remain isomorphic during MGP]\label{app:lem:isomorphic-during-execution}
Let \(G,G'\) be attributed isomorphic graphs, and let the same color filtering function \(f\) and message-passing functions \(\Theta\) be used on both. The structure-aware filtrations have the same critical values. Moreover, every isomorphism \(\sigma:G\to G'\) restricts to an isomorphism between corresponding filtration subgraphs, and \(h_v^{(\ell)}=h_{\sigma(v)}'^{(\ell)}\) at every step \(\ell\geq b_v\).
\end{lemma}

\begin{proof}
Let \(\sigma:G\to G'\) be an attributed-graph isomorphism. We first show that \(\sigma\) preserves the structure-aware filtrations. Corresponding vertices have the same colors, and both graphs use the same symmetric endpoint-color rule. Hence every edge \(\{u,v\}\in E\) satisfies
\[
x'_{\sigma(u),\sigma(v)}=x_{u,v},
\]
so corresponding edges have the same filtration value. To emphasize the underlying graph, write \(F_{S,G}^f\) and \(F_{S,G'}^f\) for the two induced functions. For every vertex \(v\), the isomorphism also gives a bijection
\(\mathcal N_v(G)\to\mathcal N_{\sigma(v)}(G')\). Therefore
\[
\begin{aligned}
F_{S,G'}^f(\sigma(v))
&=\min_{z\in\mathcal N_{\sigma(v)}(G')}
      f(x'_{z,\sigma(v)})\\
&=\min_{u\in\mathcal N_v(G)}f(x_{u,v})
=F_{S,G}^f(v),
\end{aligned}
\]
with the same convention \(\min\varnothing=0\) for isolated vertices. Thus the two filtrations have the same critical values and \(\sigma\) restricts, at every \(\alpha_\ell\), to an isomorphism
\[
G^{F_{S,G}^f}_{\alpha_\ell}
\cong
G'^{F_{S,G'}^f}_{\alpha_\ell}.
\]

We now prove the equality of corresponding embeddings by induction over the filtration steps. At step \(\ell\), newly born corresponding vertices are initialized equally because
\[
h_v^{(b_v-1)}=\Psi(x_v)
=\Psi(x'_{\sigma(v)})
=h_{\sigma(v)}'^{(b_v-1)}.
\]
For vertices born earlier, equality of the pre-update embeddings follows from the induction hypothesis. Hence, immediately before the update at step \(\ell\), every active vertex and its image have equal embeddings. The filtration isomorphism maps the active neighborhood of \(v\) bijectively onto that of \(\sigma(v)\), while preserving neighbor embeddings, the central embedding, and the edge color. The two multisets supplied to \(\bigoplus\) are therefore identical. Permutation invariance of \(\bigoplus\) and the use of the same update map \(\psi\) give
\[
h_v^{(\ell)}=h_{\sigma(v)}'^{(\ell)}.
\]
This completes the induction.
\end{proof}

\begin{theorem}[MGP is isomorphism invariant]\label{app:thm:isomorphism-invariant}
Let \(G,G'\) be attributed isomorphic graphs. For a fixed color filtering function \(f\) and any partially injective \(\Theta\),
\[
\operatorname{MGP}(G,f,\Theta)=\operatorname{MGP}(G',f,\Theta).
\]
\end{theorem}

\begin{proof}
The computation has two conceptually separate parts. First, all filtration-wise embeddings are computed. Lemma~\ref{app:lem:isomorphic-during-execution} shows that isomorphic graphs produce identical embedding trajectories on corresponding vertices. Second, the union--find pass uses those already computed embeddings to associate representatives with persistence events.

For the one-dimensional part, recall that
\(\operatorname{MGP}_1(G,f)=\mathcal D_1(G,F_S^f)\). Since the two filtration subgraphs are isomorphic at every critical value, their one-dimensional persistence diagrams are equal. Hence the one-dimensional summands agree.

For the zero-dimensional part, we use induction over the filtration values with the following invariant: every pair of corresponding connected components has alive representatives with identical embedding trajectories. This holds when a component is born by Lemma~\ref{app:lem:isomorphic-during-execution}. Now consider a value \(\alpha_\ell\). The edges entering at this value join some collection of pre-existing components into a larger component, and the isomorphism joins the corresponding components in \(G'\). By the induction hypothesis, the two collections of alive representatives have the same terminal embeddings, component by component. The representative rule removes the largest terminal embeddings until one remains, so the two merged components generate the same multiset of deaths.

It remains only to check that arbitrary choices among exactly tied representatives do not alter either the emitted tuples or the induction invariant. If \(h_v^{(n)}=h_w^{(n)}\), partial injectivity implies that \(v\) and \(w\) have identical symbolic computation trees. They therefore have the same birth value and satisfy \(h_v^{(j)}=h_w^{(j)}\) at every corresponding earlier step. Killing either one at \(\alpha_\ell\) produces the same tuple
\((\alpha_{b_v},\alpha_\ell,h_v^{(\ell)})\), while either possible survivor has the same complete trajectory. Hence the invariant is preserved independently of the remaining tie choices. After the final filtration value, corresponding surviving representatives also yield identical infinite tuples. Both disjoint-union summands of \(\operatorname{MGP}\) are equal, proving the claim.
\end{proof}

\begin{theorem}[MGP is strictly more expressive than \rephine]\label{app:thm:mgp-more-expressive-rephine}
The MGP descriptor family is strictly more expressive than the \rephine\ family:
\[
\mathfrak D_{\mathrm{MGP}}
\succ
\mathfrak D_{\mathrm{RePHINE}}.
\]
\end{theorem}

\begin{proof}
We first prove
\(\mathfrak D_{\mathrm{MGP}}\succeq
\mathfrak D_{\mathrm{RePHINE}}\).
Fix a \rephine\ diagram determined by
\(f:\mathcal X_E\to\mathbb R_{\geq0}\) and
\(g:\mathcal X_V\to\mathbb R\). We construct one MGP diagram, depending only on \((f,g)\), from which this \rephine\ diagram can be recovered on every graph.

The MGP definition allows the embedding space to be any linearly ordered set. For this construction, let \(\mathcal T\) be the set of finite symbolic message-passing computation trees and choose a set \(\mathcal H_{f,g}\) together with a bijective coding map
\[
C:\mathbb R\times\mathbb R\times\mathcal T\longrightarrow\mathcal H_{f,g}.
\]
We equip \(\mathcal H_{f,g}\) with a linear order through this coding map.
Order these codes so that
\(C(a,b,T)>C(a',b',T')\) whenever \(a>a'\), or whenever \(a=a'\) and \(b>b'\); any fixed linear order on \(\mathcal T\) resolves the remaining ties. Let \(\rho:\mathcal H_{f,g}\to\mathbb R\) recover the first argument of every code, i.e., \(\rho(C(a,b,T))=a\).

We now choose \(\Theta_{f,g}\) so that, from the first update onward, the state of every vertex \(v\) is encoded as
\[
C\bigl(g(x_v),-\gamma_v,T_v^{(\ell)}\bigr),
\qquad
\gamma_v=\min_{u\in\mathcal N_v(G)}f(x_{u,v}),
\]
where \(T_v^{(\ell)}\) is its computation tree through step \(\ell\). This can be implemented within the abstract MGP definition as follows. The initializer retains \(g(x_v)\) and the initial color, while the edge map retains \(f(x_{u,v})\). At the birth step of a non-isolated vertex, every active incident edge has filtration value \(\gamma_v\), so the aggregated message determines \(\gamma_v\); for an isolated vertex, the convention \(\min\varnothing=0\) gives the same quantity. The update preserves \(g(x_v)\) and \(-\gamma_v\) and injectively extends the computation-tree code at every later step. Hence \(\Theta_{f,g}\) is partially injective.

By construction, the largest terminal state first has the largest \(g(x_v)\), and among equal \(g\)-values it has the smallest \(\gamma_v\). The MGP representative rule therefore agrees with \rephine's rule; the computation-tree order only resolves ties for which \rephine\ permits an arbitrary choice.

The correspondence between the \rephine\ diagram and the structure-aware filtration (given in Section~\ref{sec:rephine-reinterpretation}) gives \(\alpha_{b_v}=\gamma_v\), and the death values agree. For a zero-dimensional MGP tuple define
\[
P_0(\alpha_{b_v},\alpha_{d_v},h_v)
=(0,\alpha_{d_v},\rho(h_v),\alpha_{b_v}),
\]
and map each one-dimensional tuple \((d,\infty)\) to \((1,d,0,0)\). Since \(\rho(h_v)=g(x_v)\) at every death step and at the terminal step,
\[
P\!\left(\operatorname{MGP}(G,f,\Theta_{f,g})\right)
=\mathcal R(G,f,g)
\]
for every attributed graph on the common color spaces. Thus every pair of graphs distinguished by a \rephine\ diagram is distinguished by the corresponding MGP diagram.

For strictness, let \(G\) be the four-vertex star and \(H\) the four-vertex path, with one common vertex color. Because edge colors depend only on endpoint colors, all edges in both graphs also have one common edge color. For every \rephine\ choice \((f,g)\), both graphs have four vertices, three edges, one connected component, and no cycles; moreover, all values \(\lambda_v=g(x_v)\) and \(\gamma_v=f(x_{u,v})\) coincide. Their \rephine\ diagrams therefore contain the same three finite tuples and the same one infinite tuple.

In the structure-aware filtration, all vertices and edges appear at a single filtration step. Partial injectivity makes the resulting vertex state distinguish different neighbor-color multisets and hence different degrees. Every vertex contributes exactly one zero-dimensional tuple, so the tuple-embedding multiset contains the degree multiset. These are
\(\{\!\!\{3,1,1,1\}\!\!\}\) for the star and
\(\{\!\!\{2,2,1,1\}\!\!\}\) for the path. Consequently every descriptor in \(\mathfrak D_{\mathrm{MGP}}\) distinguishes \(G\) and \(H\), whereas no descriptor in \(\mathfrak D_{\mathrm{RePHINE}}\) does. Hence
\[
\mathfrak D_{\mathrm{MGP}}
\succ
\mathfrak D_{\mathrm{RePHINE}}.
\]
\end{proof}

\section{Dataset details}
\label{app:dataset-details}

\begin{table}[h]
\centering
\newtable
\caption{Datasets: number of graphs, mean numbers of vertices and undirected edges, task and majority-class rate (classification). Graphs without node features receive one-hot encoded degrees.}
\label{tab:datasets}
\footnotesize
\resizebox{\linewidth}{!}{%
\begin{tabular}{llrrrlc}
\toprule
\textbf{Dataset} & \textbf{Graphs of} & \textbf{Graphs} & \(\bar n\) & \(\bar e\) & \textbf{Task} & \textbf{Majority} \\
\midrule
MUTAG       & nitro compounds, mutagenicity         & 188    & 17.93 & 19.79 & 2 classes & 0.6649 \\
PTC\_MR     & compounds, carcinogenicity (male rats) & 344 & 14.29 & 14.69 & 2 classes & 0.5581 \\
PROTEINS    & proteins, enzyme or not               & 1,113  & 39.06 & 72.82 & 2 classes & 0.5957 \\
NCI1        & compounds, lung-cancer screen         & 4,110  & 29.87 & 32.30 & 2 classes & 0.5005 \\
NCI109      & compounds, ovarian-cancer screen      & 4,127  & 29.68 & 32.13 & 2 classes & 0.5038 \\
ZINC (12k)  & molecules, constrained solubility     & 12,000 & 23.2  & 24.9  & regression, MAE & -- \\
\bottomrule
\end{tabular}}
\end{table}

\new{ZINC uses its public split of 10,000 training, 1,000 validation and 1,000 test graphs.}

\section{Training and selected configurations}
\label{app:training}

\begin{newpart}
\noindent\textbf{Training (Table~\ref{tab:main}).} We train all models using Adam with learning rate \(10^{-3}\), weight decay \(10^{-5}\), batch size 64 (32 for MGP on PROTEINS) and at most 300 epochs. We use early stopping with patience 30 on the validation score (60 for MGP on NCI109; on NCI1, MGP uses stochastic weight averaging from epoch 150 without early stopping). The shared backbones and selected MGP configurations are given in Table~\ref{tab:selected-b}, with the MGP instances described in Appendix~\ref{sec:instances}.

The MGP hyperparameter search includes \(k\), \(F\), \(T\), the number of message-passing rounds, aggregation, update depth and width, normalization, residual connections, step embeddings, trajectory summaries, parallel schedules, read-out, read-out learning-rate multiplier and dropout. For NCI1, NCI109 and ZINC, we also search over GatedGCN, feed-forward, structural-encoding and component-virtual-node options. For every dataset, we perform a final search over combinations of component attention, substructure counts and cycle lengths, considering seven combinations. We also consider three read-out averaging and training variants: exponential moving average, early-stopping patience 60 and stochastic weight averaging \citep{izmailov2018swa}. For SpectRe, we use one eigenvalue per event, computed by power iteration at every third event of the filtration.

\begin{table}[h]
\centering
\caption{Selected configurations for Table~\ref{tab:main}. Backbone (all methods): GINE layers, width, jumping knowledge (JK), RWSE. MGP: aggregation (Appendix~\ref{sec:instances}), rounds per step \(R\), \(F\) and \(k\); \(T=4\) and \(T=16\) run in parallel.}
\label{tab:selected-b}
\tablesize
\begin{tabular}{lllcc}
\toprule
\textbf{Dataset} & \textbf{Backbone} & \textbf{MGP} & \(F\) & \(k\) \\
\midrule
MUTAG    & 5 layers, 256, JK       & GINE-\(\varepsilon\), \(R=3\)   & 16 & 64 \\
PTC\_MR  & 5 layers, 256, JK       & GINE-\(\varepsilon\), \(R=3\)   & 16 & 64 \\
PROTEINS & 5 layers, 128, JK, RWSE & GatedGCN + attention           & 8  & 64 \\
NCI109   & 5 layers, 256, RWSE     & GINE-\(\varepsilon\), \(R=2\)   & 16 & 32 \\
NCI1     & 3 layers, 256, JK       & GatedGCN + attention, \(R=2\)  & 8  & 32 \\
ZINC     & 3 layers, 128, JK, RWSE & GatedGCN                       & 16 & 128 \\
\bottomrule
\end{tabular}
\end{table}
\end{newpart}

\section{BREC}
\label{app:brec}

BREC \citep{wang2024brec} consists of 400 pairs of non-isomorphic graphs (800 total graphs), organized into four categories: Basic, Regular, Extension and CFI. Basic graphs consist of 60 pairs of small graphs that are indistinguishable by 1-WL, exposing the simplest failures of classical GNN architectures. Regular graphs consist of 140 pairs of graphs further divided into simple regular graphs, strongly regular graphs, 4-vertex condition regular graphs and distance regular graphs. For further details about the remaining graph structures, we refer to \citet{wang2024brec}.

\begin{newpart}
For the experiments in Tables~\ref{tab:brec-results} and~\ref{tab:brec-full}, MGP uses the selected vertex statistic as node colors and the curvature values for the structure-aware edge filtration. We use the distinct-value schedule (\(T=\infty\)), one filtration, GINE-type aggregation and randomly initialized parameters \(\Theta\) with \(k=8\). We report the mean over three seeds.

We compute all descriptors in double precision and convert them to canonical form. Two graphs are considered distinguished when their canonical forms differ by more than \(10^{-6}\) in at least one entry. Following \citet{wang2024brec}, we also perform a reliability check. Each graph is evaluated under three random relabelings, including changes in vertex labels, vertex order, edge order and direction. A graph is considered reliable when all relabelings produce the same descriptor as the original graph. We count a pair as distinguished only when both graphs satisfy this condition.
\end{newpart}

\begin{table}[t]
\centering
\newtable
\caption{Accuracy on the BREC categories by vertex filtration function and diagram type, under the reliability check. Regular: all 140 regular pairs; every distinguished one is simple regular (strongly, 4-vertex condition and distance regular: 0.00 for all). MGP: mean over three seeds.}
\label{tab:brec-full}
\vspace{4pt}
\tablesize
\begin{tabular}{llccccc}
\toprule
\textbf{Filtration} & \textbf{Diagram} & \textbf{CFI} & \textbf{Regular} & \textbf{Extension} & \textbf{Basic} & \textbf{Total} \\
\midrule
\multirow{4}{*}{Degree}
  & \(\mathrm{PH}^{V}\) & 0.03 & 0.00 & 0.07 & 0.03 & 0.03 \\
  & \(\mathrm{PH}^{E}\) & 0.03 & 0.34 & 0.55 & 0.98 & 0.41 \\
  & RePHINE              & 0.03 & 0.34 & 0.55 & 0.98 & 0.41 \\
  & MGP                  & 0.03 & 0.34 & 0.54 & 0.98 & 0.41 \\
  \cmidrule(lr){1-7}
\multirow{4}{*}{Betweenness}
  & \(\mathrm{PH}^{V}\) & 0.10 & 0.34 & 0.82 & 0.95 & 0.49 \\
  & \(\mathrm{PH}^{E}\) & 0.03 & 0.34 & 0.55 & 0.98 & 0.41 \\
  & RePHINE              & 0.09 & 0.35 & 0.91 & 1.00 & 0.52 \\
  & MGP                  & 0.10 & 0.35 & 0.91 & 1.00 & 0.53 \\
  \cmidrule(lr){1-7}
\multirow{4}{*}{Closeness}
  & \(\mathrm{PH}^{V}\) & 0.10 & 0.09 & 0.46 & 0.27 & 0.21 \\
  & \(\mathrm{PH}^{E}\) & 0.03 & 0.34 & 0.55 & 0.98 & 0.41 \\
  & RePHINE              & 0.10 & 0.35 & 0.82 & 1.00 & 0.50 \\
  & MGP                  & 0.09 & 0.35 & 0.80 & 1.00 & 0.49 \\
  \cmidrule(lr){1-7}
\multirow{4}{*}{PageRank}
  & \(\mathrm{PH}^{V}\) & 0.03 & 0.00 & 0.09 & 0.05 & 0.04 \\
  & \(\mathrm{PH}^{E}\) & 0.03 & 0.34 & 0.55 & 0.98 & 0.41 \\
  & RePHINE              & 0.03 & 0.34 & 0.55 & 0.98 & 0.41 \\
  & MGP                  & 0.03 & 0.34 & 0.55 & 0.98 & 0.41 \\
\bottomrule
\end{tabular}
\end{table}

For all vertex filtering choices, \(\mathrm{PH}^E\) remains constant because it depends only on the edge filtration. Conversely, \(\mathrm{PH}^V\) does not use Forman--Ricci curvature. We make some remarks about degree in particular, as it is the most canonical vertex filtering value on an unlabeled graph. \(\mathrm{PH}^V\) is unable to distinguish almost any graph pairs, confirming that degree information alone is not sufficiently discriminative. \(\mathrm{PH}^E\), whose induced filtration places all vertices at zero, yields significant improvements, while RePHINE coincides with \(\mathrm{PH}^E\) in this setting. RePHINE improves over or matches \(\mathrm{PH}^E\), and, except for Betweenness, \(\mathrm{PH}^E\) outperforms \(\mathrm{PH}^V\).

\begin{newpart}
\section{Message functions used in practice}
\label{sec:instances}
Definition~\ref{def:mgp} specifies the general form of \(\Theta\), consisting of an initializer \(\Psi\), an edge map \(\phi\), a permutation-invariant aggregation \(\bigoplus\) over the active neighborhood, and an update function \(\psi\). In the experiments, \(\Psi\), \(\phi\) and \(\psi\) are MLPs, and we use \(\mathcal{H}=\mathbb{R}^k\) with lexicographic ordering. Let \(G_\ell\) denote the filtration subgraph at step \(\ell\). We consider the following aggregation functions, selecting one for each dataset using the validation score:
\begin{itemize}[itemsep=1pt,topsep=2pt]
\item GINE \citep{hu2020strategies}: \(h_v^{(\ell)}=\psi\bigl(h_v^{(\ell-1)},\sum_{u\in\mathcal{N}_v(G_\ell)}(h_u^{(\ell-1)}+\phi(x_{u,v}))\bigr)\);
\item GINE-\(\varepsilon\): the same sum plus \((1+\varepsilon)\,h_v^{(\ell-1)}\), where \(\varepsilon\) is a learned scalar \citep{gin};
\item GAT \citep{DBLP:conf/iclr/VelickovicCCRLB18}: multi-head attention over the active incident edges of \(v\), with scores computed from \((h_u^{(\ell-1)},h_v^{(\ell-1)},\phi(x_{u,v}))\);
\item a graph-transformer layer \citep{dwivedi2020generalization} whose attention is restricted to the active neighbors, with keys and values \(Wh_u^{(\ell-1)}+\phi(x_{u,v})\);
\item GatedGCN \citep{bresson2017residual}, which keeps a state \(e_{u,v}\) on every active edge, initialized to \(\phi(x_{u,v})\) when the edge enters and updated at every round;
\item GatedGCN + attention, a GPS layer \citep{rampasek2022recipe} in which GatedGCN is followed by multi-head attention among the active vertices of the same connected component of \(G_\ell\).
\end{itemize}
The first four are permutation-invariant functions of the multiset of triples \((h_u^{(\ell-1)},h_v^{(\ell-1)},\phi(x_{u,v}))\) over the active neighbors. Therefore, they satisfy Definition~\ref{def:mgp}.

We also search over \(R\) message-passing rounds per step, using separate weights for each round; a sinusoidal embedding of \(\ell/T\), added to the aggregated message; and a component virtual node, which adds \(W\operatorname{mean}_{w\in C_\ell(v)}h_w^{(\ell-1)}\) to the message, where \(C_\ell(v)\) is the connected component of \(v\) in \(G_\ell\).

Other options include random-walk structural encodings of \(G\) \citep{dwivedi2022lspe}, counts of cycles of length 3 to 8 through \(v\) in \(G\) \citep{bouritsas2023gsn}, which are added to \(\Psi(x_v)\), and the length of the shortest cycle of \(G\) through the edge responsible for each one-dimensional tuple, which is appended to that tuple. We also consider a residual feed-forward block after each round \citep{luo2025classic}.

For zero-dimensional tuples, we optionally replace \(h_v\) with the trajectory summary
\[
\Bigl(h_v^{(b_v)},\ \frac{1}{d_v-b_v+1}\sum_{\ell=b_v}^{d_v}h_v^{(\ell)},\ h_v^{(d_v)}\Bigr).
\]
We also consider running two step schedules, \(T=4\) and \(T=16\), in parallel, each with its own read-out.

At each step, these functions depend on the multiset of active neighbor states, the connected component of the vertex in \(G_\ell\), and isomorphism-invariant vertex and edge encodings. The induction argument used in the proof of Lemma~\ref{lem:isomorphic-during-execution} also applies to these functions. For GatedGCN, the induction considers vertex and edge states jointly. For the component virtual node, it considers the connected components of corresponding filtration subgraphs, which are preserved by the isomorphism. The trajectory summaries also agree because they are computed from corresponding embedding trajectories. Hence, Theorem~\ref{thm:isomorphism-invariant} extends to these instances under the corresponding partial-injectivity assumption. Theorem~\ref{thm:mgp-more-expressive-rephine} concerns the family of Definition~\ref{def:mgp} and is not affected.
\end{newpart}

\section{Implementation}
\label{app:runtime}

\begin{algorithm}[!ht]
\caption{High-Level MGP Diagram Computation}
\label{alg:mgp-algo}
\begin{algorithmic}[1]
\Require \(G=(V,E,x)\), structure-aware filtration \(F_S^f\), \(\Theta=(\Psi,\psi,\phi,\bigoplus)\)
\Ensure \(\operatorname{MGP}_0,\operatorname{MGP}_1\)
\State \(\operatorname{MGP}_0\gets\emptyset\), \(\operatorname{MGP}_1\gets\emptyset\), and \(G_0\gets\emptyset\)
\State Let \(\alpha_1<\cdots<\alpha_n\) be the critical values of \(F_S^f\)
\Statex \textbf{Step 1: compute embeddings}
\For{\(\ell=1\) to \(n\)}
  \State Form \(G_\ell\) from \(G_{\ell-1}\) by adding all simplices with value \(\alpha_\ell\)
  \ForAll{vertices \(v\) born at \(\ell\)}
    \State \(h_v^{(b_v-1)}\gets\Psi(x_v)\)
  \EndFor
  \ForAll{\(v\in V(G_\ell)\)}
    \State compute \(h_v^{(\ell)}\) using Definition~\ref{def:mgp}
  \EndFor
\EndFor
\Statex \textbf{Step 2: track births and deaths}
\State Initialize Union--Find with all vertices marked inactive
\For{filtration batches \(\alpha_\ell\) in increasing order}
  \State activate vertices born at \(\alpha_\ell\)
  \ForAll{edges \(e=\{u,v\}\) with \(f(x_{u,v})=\alpha_\ell\)}
    \If{\(u\) and \(v\) lie in distinct components}
      \State kill the representative \(w\) with the largest \(h_w^{(n)}\)
      \State \(\operatorname{MGP}_0\gets\operatorname{MGP}_0\mathbin{\dot\cup}\{(\alpha_{b_w},\alpha_\ell,h_w^{(\ell)})\}\)
      \State merge the two components and retain the other representative
    \Else
      \State \(\operatorname{MGP}_1\gets\operatorname{MGP}_1\mathbin{\dot\cup}\{(\alpha_\ell,\infty)\}\)
    \EndIf
  \EndFor
\EndFor
\Statex \textbf{Step 3: record survivors}
\ForAll{remaining representatives \(v\)}
  \State \(\operatorname{MGP}_0\gets\operatorname{MGP}_0\mathbin{\dot\cup}\{(\alpha_{b_v},\infty,h_v^{(n)})\}\)
\EndFor
\State \Return \(\operatorname{MGP}_0\sqcup\operatorname{MGP}_1\)
\end{algorithmic}
\end{algorithm}

\begin{newpart}
We define the priority of each vertex by the rank of its terminal embedding, using lexicographic ordering and breaking exact ties by the lower vertex index. The representative of a component is the vertex with the lowest priority. A vertex dies at the first step in which its component contains a vertex of lower priority. We compute these events using a sequential union--find algorithm with complexity \(O(\abs{E}\,\alpha(\abs{V}))\), where \(\alpha\) is the inverse Ackermann function.

For small graphs, we use a dense implementation that computes the same information for all filtration steps simultaneously. The adjacency matrices of the filtration subgraphs in a batch are transitively closed using \(\lceil\log_2 \abs{V}\rceil\) batched matrix products, without host synchronization. For larger graphs, we use label propagation to compute the connected components at each filtration step on the device.

\noindent\textbf{SpectRe.} The official SpectRe kernel runs on the CPU and performs an eigendecomposition for every merged component, requiring 1078\,s per epoch on NCI1. We implemented a GPU version that computes component eigenvalues in batches. Its persistence tuples agree with the official kernel on 600 graph--filtration pairs.

Its fast variant follows the fast kernel of the official implementation, retaining only the largest eigenvalue for components with at least 10 vertices, computed using batched power iteration. For Table~\ref{tab:main}, we use a reduced version of the fast variant, computing one eigenvalue for each component with at least two vertices at every third event of the filtration. Since events are ordered by filtration value, selecting events by position may depend on the ordering of tied values. Thus, this setting is not exactly permutation invariant when ties occur.

\end{newpart}

\end{document}